\documentclass[letterpaper, 10 pt, conference]{ieeeconf}  

\IEEEoverridecommandlockouts                              

\usepackage{graphicx}
\usepackage{xcolor}
\usepackage{amsmath} 
\usepackage{placeins}  
\usepackage{calc}   
\usepackage{caption}
\usepackage{subcaption} 
\usepackage{lastpage}
\usepackage{multirow}
\usepackage{hyperref}
\usepackage{flushend}
\usepackage[normalem]{ulem}

\renewcommand{\arraystretch}{1.3}

\newcommand{\brox}[1]{\textcolor{purple}{#1}}

\usepackage{xspace}

\newcommand{\ie}{\mbox{i.\,e.}\xspace}

\renewcommand{\[}{\begin{equation}}
\renewcommand{\]}{\end{equation}}

\renewcommand{\eqref}[1]{Eq.~(\ref{#1})}

\newcommand{\robotio}{RobotIO\xspace}

\usepackage{rotating}
\usepackage{accents}
\usepackage{amsfonts}
\usepackage{pifont}
\newcommand{\xmark}{\ding{55}}%
\newcommand{\tmark}{$\sim$}
\newcommand{\cmark}{\checkmark}
\newcommand{\myvec}[1]{\accentset{\rightharpoonup}{{#1}}}
\newcommand{\Max}[1]{\textcolor{red}{#1}}
\newcommand{\Evan}[1]{\textcolor{blue}{#1}}

\title{\LARGE \bf
\robotio: A Python Library for Robot Manipulation Experiments}

\author{Lukas Hermann$^*$, Max Argus$^*$, Adrian Röfer, Abhinav Valada, Thomas Brox
\thanks{* equal contribution authors}
\thanks{All authors are at the University of Freiburg and are members of Brain-Links-Brain-Tools. The corresponding author can be contacted via: {\tt\small argusm@cs.uni-freiburg.de}}%
}

\begin{document}

\maketitle
\thispagestyle{empty}
\pagestyle{empty}

\begin{abstract}


Setting up robot environments to quickly test newly developed algorithms is still a difficult and time consuming process. This presents a significant hurdle to researchers interested in performing real-world robotic experiments.
\mbox{\robotio} is a python library designed to solve this problem. It focuses on providing common, simple, and well structured python interfaces for robots, grippers, and cameras, etc. 
These are provided with implementations of these interfaces for common hardware.
This enables code using \robotio to be portable across different robot setups.
In terms of architecture, \robotio is designed to be compatible with OpenAI gym environments, as well as ROS; examples of both of these are provided.
The library comes together with a number of helpful tools, such as camera calibration scripts and episode recording functionality that further support algorithm development.
\end{abstract}

\vspace{1em}
\section{Introduction}



Recently there have been rapid developments in robotics, machine learning, and computer vision.
This has led to an increase in interest in the field and an increased motivation to conduct real-world robotic experiments. However, setting up such an environment remains a challenging task, especially for researchers who do not come from a robotics background.

Despite a large amount of work being performed in simulation, real robotic experiments remain an important test of any algorithm's practical viability. To enable researchers to achieve this goal we present the \robotio library, which is designed to let new developers hit the ground running and to get new algorithms working on a robot as quickly as possible. The library provides hardware independent interfaces allowing for portable code to be developed and to be shared between robots and labs.

The library focuses on providing simple, and well structured interfaces for controlling robots. \robotio focuses on co-bots that are commonly used for research, such as the Kuka iiwa, the Universal-Robots UR3, and the Franka-Emika Panda. In addition, interface code for common cameras, such as the Intel RealSense or the Microsoft Kinect are provided.

This document aims to describe the problems addressed by the \robotio library so that potential users can evaluate if it is a good fit for their tasks. The modular structure of the library also allows it to be extended easily to new setups, this document will also explain the library's design choices in order to help developers interested in extending it for their purposes.
    
\begin{figure}[!htb]
\centering
\includegraphics[width=.5\textwidth]{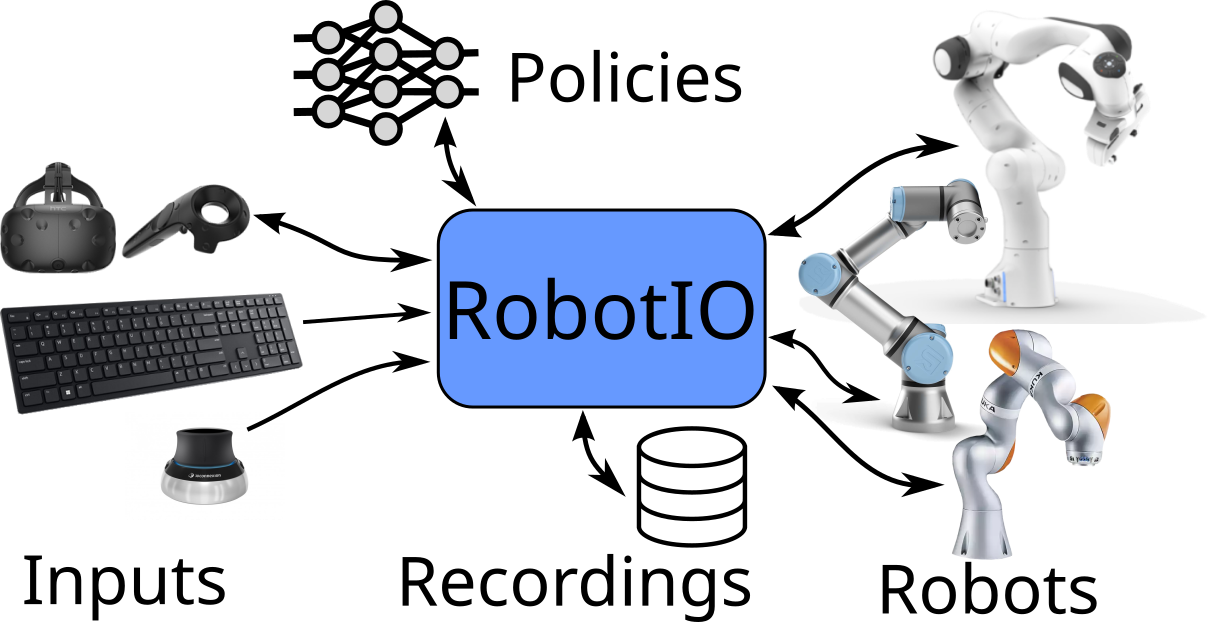} 
\caption{\robotio provides shared interface classes to quickly interface between robots and input devices. This allows for the quick setup of real-world robot experiments.}
\label{fig:teaser}
\end{figure}

\section{Background}
\label{sec:related_work}




Robotic systems feature software stacks with various modules solving specific functions.
To provide a better overview of this field we will summarize some of the common functionality here.

\textbf{Robot Controllers}: When controlling a robot one wants to have an interface that gives as fine grained control as necessary and is as abstract as possible. In order to be easily usable, robots are normally shipped with programming functionality. Historically this has been through vendor-specific domain languages, recently this has shifted to robots being provided with control APIs in real programming languages such as Java~\cite{kuka2016osmanual}, C++~\cite{libfranka}, and Python, with current robotics research settling on Python as the most common language used for development. Robot controllers typically allow inputting joint angles or Cartesian poses that a robot should move to. If a controller API in a common language is not provided by the manufacturer, these are often programmed by researchers and available online~\cite{ur-rdte}.

The PyRobot~\cite{Murali2019} project aims to provide a simple, easy-to-use Python interface, similar to our work. However, this work is focused on the locobot platform and has not been extended to multiple robots.


\textbf{Motion Libraries}: 
Typically, there are a number of standard goals that users have for their robots, such as moving to a desired end-effector pose from a starting pose. Motion libraries generate trajectories for these types of goals while enforcing some additional constraints such as straight-line Cartesian motion, low accelerations, and jerks. In addition, these libraries typically include an execution component that executes the planned trajectories, while closely monitoring the state of the execution. Examples of these include Reflexxes~\cite{Kroeger2011} and frankx~\cite{Berscheid2021}. Notably, these libraries do not address the problem of avoiding obstacles. They are solely concerned with generating primitive motions with desirable dynamic properties.

\textbf{Planning Libraries}: 
In many cases, robots operate in more cluttered environments. In such environments, a motion generator is prone to generating motions that are in collision. Motion planning libraries are thus concerned with generating trajectories to satisfy a given goal, while, given a model of the environment's geometry, avoiding contact with the surroundings. The most well-known example is Moveit~\cite{Coleman14}, built on OMPL~\cite{sucan2012ompl, kingston2019exploring} and robowflex~\cite{Kingston2021} as a simplification. Alternatives are KOMO~\cite{Toussaint2014newton, toussaint2015logic}, and Traj-Opt~\cite{schulman2014motion}. Both are less feature-complete than MoveIt but offer more options in terms of goals and constraints for the motion.


\textbf{Robot Simulators}: Simulation is an important part of robot algorithm development. Among the most common ones are OpenRAVE~\cite{Diankov2008}, Gazebo~\cite{Koenig2004} the standard simulation tool of the ROS ecosystem, PyBullet~\cite{Coumans2021} a lightweight physics solver with generation tools for sensor data, and Mujoco~\cite{Todorov2012} a physics solver similar to PyBullet in its feature-set, but with the additional ability of contact differentiability. The latter has become an interesting feature for back-propagation in deep machine learning models, and as such, has become a desired feature for the next generation of simulators such as Brax~\cite{brax2021github}. However, so far a new standard simulator has not emerged. In addition to the open-source simulators, there are also some proprietary options, usually offering much more realistic renderings of camera data, easing the transfer to the real world. Notable are Unity3d's efforts to create a simulation ecosystem for robotics~\cite{unity_robotics}, and Nvidia's ISAAC simulation environment~\cite{Makoviychuk2021}, boasting both their physics solver and raytracing-enabled real-time rendering.

\textbf{Simulated Robot Environments}:
%
While \robotio focuses on enabling real-world robot experiments, for completeness we briefly include a summary of simulated robot environments. OpenAI gym~\cite{Brockman2016} and DeepMind Control-Suite~\cite{Tassa2018} provide test environments for continuous control, especially for RL algorithms. More realistic simulated environments include \cite{Szot2021, Kolve2017} these combine realistic 3D environments with navigation and manipulation tasks.

\textbf{Robotics Frameworks}:
Robotics frameworks seek to provide an environment and tools for the complete development of robotic applications. While there are less well-known ones such as Microsoft Robotic Studio \cite{microsoft_robotic_studio}, and \emph{Rock}~\cite{dfkirock}, the Robot Operating System (ROS)~\cite{Quigley2009} is the de facto standard framework in robotics today. It provides an ecosystem of different modules that can be tied together by a multi-device communication protocol. Additionally, it ships with simulation, visualization tools and debugging tools for analyzing the communication of different modules.
A large number of packages for robotics problems, such as control, path-planning, or image processing, are available in ROS. While \robotio does provide some of the functionality provided by these packages, we do want to stress that it is not meant to rival the ROS ecosystem. Instead, we see \robotio as a complementary library into which ROS modules can easily be integrated, as shown in (ref Franka interface), avoiding the need for inexperienced users to interface with all the different components of a typical ROS system manually.


\begin{figure*}[!t]
\centering
 \centering
 \includegraphics[width=.65\linewidth]{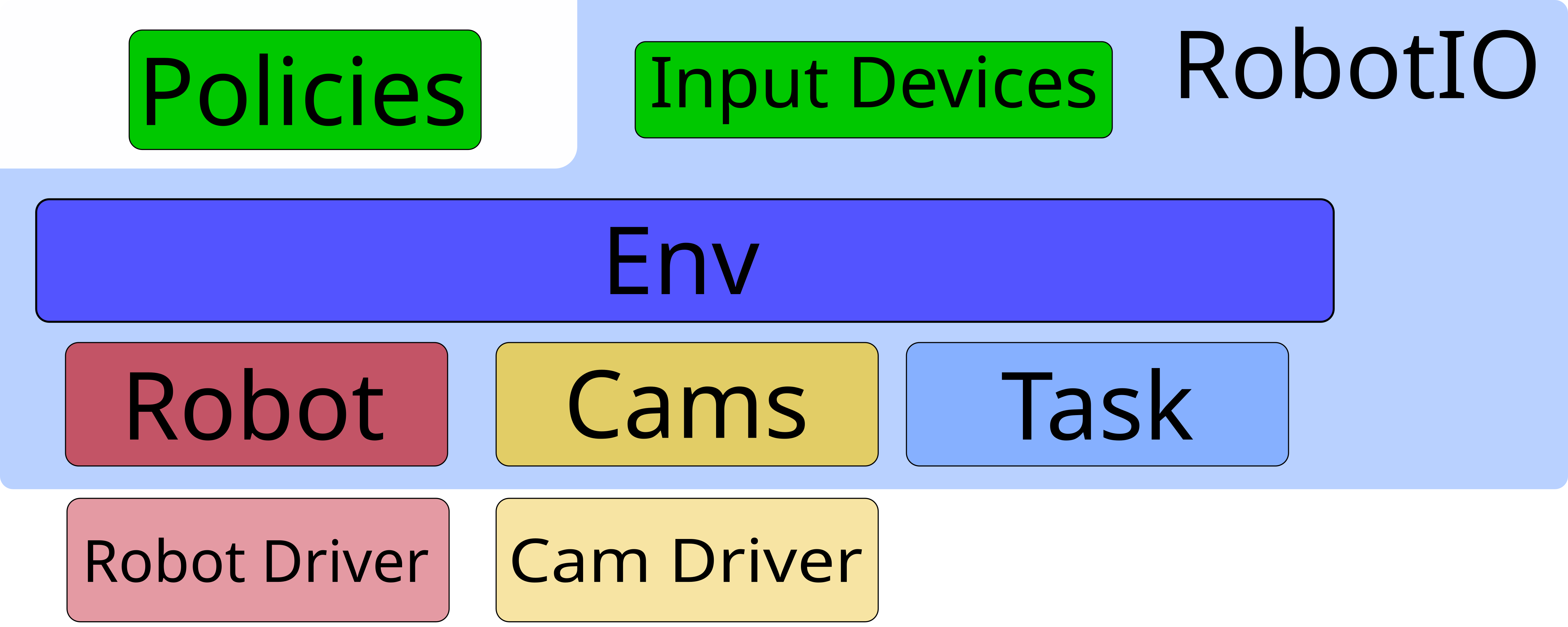}
 \caption{A high-level overview of \robotio's architecture, showing the composition of an environment through various modules. }
 \label{fig:arch}
\end{figure*}

\vspace{1em}
\section{The \robotio Library}
\label{sec:method}

\robotio is designed as a toolbox, which enables easy access to both robots and common sensors so that anyone can use these physical devices in their experiments. As a more structured approach to robotic experiments, \robotio also provides a gym-like environment, which makes use of only a few functions for stepping and resetting the setup. Instantiating a setup in this way allows easy passing of actions to the robot, often without having to consider the complexity of the underlying setup.
Individual components of \robotio can be run independently, with examples of this usually given in the relevant class files.

\subsection{Architecture Overview}


The architecture of the \robotio library is described in Figure \ref{fig:arch}.
We start by briefly summarizing the main modules that environments are composed of, these are described in more detail in the following sections.

\textbf{InputDevice/Algorithm}: At the top level of the control stack the input for the environment is generated. This can come from a human input device, such as a keyboard or 3D mouse, or from code, such as an ML policy. The latter usually consumes environment states as inputs in order to compute current actions.

\textbf{Environment}: \robotio is designed to provide simple gym-like environments, which make use of only a few functions for stepping and resetting the setup. Instantiating a setup in this way allows easy passing of actions to the robot, often without having to consider the complexity of the underlying setup. 

\textbf{Robots}: The robot interface handles the control of a single robot arm. The purpose of this class is to translate actions into robot motion, by calling \textbf{robot controller} code.
\footnote{As robot controller could have multiple interpretations, for this document we use it to refer to the code that takes actions and moves the robot.}

\subsection{Motion Control Actions}


Having interchangeable environments and robots requires having standardized actions. Default gym design has vector actions, with the meaning of the action information specified as part of the environment. As it can be useful to use different action types during one session we have instead chosen dictionary actions that contain the meaning of the action explicitly. Other system parameters, such as the TCP position are part of the configuration.

The most common action type is Cartesian position control. In \robotio these are encoded as follows.

\begin{verbatim}
action = {motion=((x,y,z),(x,y,z,w),g),
          ref=abs/rel,
          path=ptp/lin,
          blocking=True/False}
\end{verbatim}

Gripper actions, $g$, are discretized by default $g \in (-1,1)$ for closing and opening respectively, we have found this to work well in practice. Following standard robotics practice, all action sizes are given in meters, this also applies to simulation shape sizes.

\subsection{Input Devices}


Common challenges in robotics are checking the environment, and, especially in robotic learning, collecting demonstrations. Typically, this is quite cumbersome, as collecting data from external input devices is not a standard feature of Python.
With \robotio we provide simple implementations to a number of common input devices such as keyboards, 3D mice, and even VR controllers. These implementations communicate asynchronously with the respective device and the retrieved input can then be read and processed by the main application. This allows users to easily control their robots manually in order to explore their environment, check safety margins and the general response of the robot's control, and collect the aforementioned demonstration trajectories as input for learning algorithms.

While all the different interfaces are easy to use, we do feel it necessary to point out that their control spaces are not identical: Keyboards and 3D mice output relative actions, while VR controllers can give absolute positions. When using absolute positions additional filtering to guarantee safe operation, especially as VR controllers can accidentally be occluded during operation.

In the background, \robotio uses PyBullet's VR interface to connects to the devices which connects to SteamVR. This means that it is possible to connect to any mainstream VR headset, be it a Valve Index, HTC Vive, or one of the Occulus devices. 

\begin{verbatim}
input = KeyboardInput(**cfg)
while True:
    action = input.get_action()
    robot.move_cart_pos(action)
\end{verbatim}

\subsection{Robots}


The robot takes input actions and moves the robot and instigates the motion of the robot by accessing the software interface provided by the robot. In the initial release of RobotIO we provide code for the following robots: Kuka iiwa, Universal Robots, and Franka-Emika Panda. This functionality relies on the robot software provided by the vendors or makes use of open-source third-party libraries that wrap these. Writing interfaces for new robots should be easy if these already have decent APIs. The panda{\_}frankx class provides an example of wrapping a C++ interface with Cython. Some notes on system requirements of robot interfaces are given in Table~\ref{tab:modules}. 

In addition, \robotio provides a ROS implementation of its robot interface, which wraps any implementation of the robot interface in a ROS node and adds the functionality of automatically publishing status information to the ROS network.

\begin{verbatim}
robot = RobotClass(**cfg)
robot.move_to_neutral()
robot.move_cart_pos(action)
robot_state = robot.get_state()
\end{verbatim}

\subsection{Grippers}

Most robots do not come with integrated grippers, for this reason, an interface to grippers is also required. The gripper interface is usually contained within the robot. In this initial release, \robotio provides gripper interfaces for the Weiss/Schunk WSG-50 gripper as well as the Weiss Griplink system, which has been tested using the CRG-30 gripper. The interface to the Franka-Emika gripper is handled through the panda\_frankx library.


\begin{verbatim}
gripper = GripperClass(**cfg)
gripper.move_to(action)
gripper_state = gripper.get_state()
\end{verbatim}

\begin{figure*}[!t]
 \centering
 \includegraphics[width=.32\linewidth]{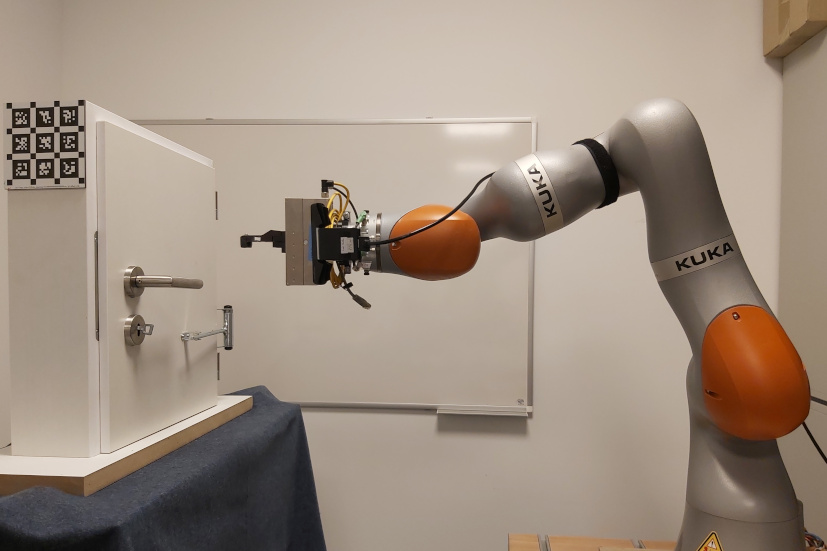}
 \includegraphics[width=.32\linewidth]{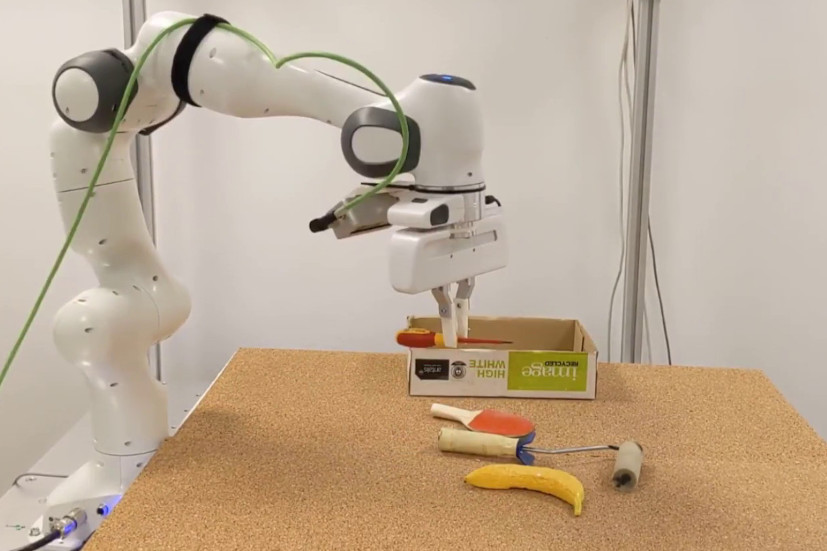}
 \includegraphics[width=.32\linewidth]{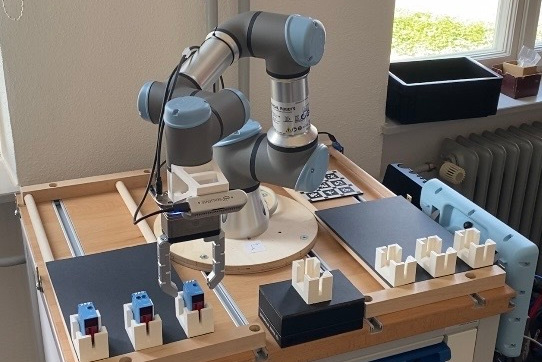}
 \caption{Examples of robot setups that have been used together with \robotio. Left: Setup for opening a door using a KUKA iiwa used in~\cite{Nematoli2020}; Middle: Pick and place using a Franka-Emika Panda~\cite{Borja2022}; Right: UR-3 robot operated through \robotio.}
 \label{fig:examples}
\end{figure*}

\subsection{Cameras}


Another common task, especially for researchers in computer vision, is the integration of cameras into their pipeline. Outside of the ROS-context, this can be a very cumbersome and time consuming problem. Thus, \robotio provides a simple camera interface and a multi-threaded wrapper that fetches the most recent data from cameras implemented under this interface. The camera interface also requires the implementation of common utility functions such as projecting and de-projecting points, generating camera and projection matrices, and computing point clouds.

Under this interface, \robotio has currently implemented connections for two cameras which are very common in robotics research: The Intel RealSense and the Microsoft Azure.

In today's robotics research it has become standard practice to use more than one camera in a robotic system, \ie typically a static camera observes the scene while a wrist-mounted camera observes the interactions of the gripper in greater detail.
To manage multiple cameras, \robotio provides a camera manager also handles additional tasks such as calibration, serialization, and depth normalization. The \textbf{camera calibration} functionality is able to perform end-of-arm camera calibration, as well as base-to-camera calibration, by employing an AprilTag marker detection system~\cite{Olson2011, Freipose2020}. 
\begin{verbatim}
cam = CameraClass(**cfg)
rgb, depth = cam.get_image()

\end{verbatim}

\subsection{Tactile Sensors}

In recent years, high-resolution tactile sensors \cite{yuan2017gelsight, Lambeta2020, bhirangi2021reskin} have become more popular in robotic manipulation research, as they are able to retrieve shape information and very small forces in a given grasp, which enables high-fidelity manipulation. Aligned with this trend, \robotio also provides an interface to one of them: The low-cost digit tactile sensor~\cite{Lambeta2020}. Much like in the case of the cameras, the implementation allows for synchronous and asynchronous access.

\subsection{Trajectory Collection Tools}


A number of often-needed tools are also provided. The first of these is a \textbf{data recorder} which saves environment steps to disk. The initial release contains two different data loaders, a simple recorder with minimal functionality as well as a VR recorder which includes helpful functions to be able to control episode saving with VR controls. In addition to this recorder, a \texttt{PlaybackEnv} class is provided which de-serializes recordings and emulates the interface of a live environment. This avoids code duplication which would occur when reading state information from observation dictionaries.

\subsection{Safety, Workspaces, and Motion Controllers}

Personal safety is a major concern when working with robotic systems, especially when code is being developed. When testing code it is strongly advised to do this with low maximum velocities which can be configured in the robot controller configurations. 
\footnote{please note that the ur-rdte controller does not make use of the maximum velocity parameter that can be specified when executing servo motions.}

In addition to this, \robotio provides \textbf{workspaces} that automatically clip the range of motions that a robot can navigate to. Currently, these are limited to axis-aligned boxes in the frame of the robot's base.

To ensure greater safety in human operations, \robotio ships with an action-filtering class called \texttt{RelActionControl}. This class turns relative actions into absolute ones, while also limiting exerted forces, and preventing high jerks, accelerations, or dangerous contortions.

\subsection{Simulated Environments}


While the \robotio library primarily focuses on interface specification for hardware, having simulations that are compatible with \robotio is very helpful for research and development.
While these are not integrated into the library, we would like to point to the CALVIN~\cite{Mees2022} environment\footnote{CALVIN environment: \url{https://github.com/mees/calvin}}, which focuses on learning from play data that can be collected in simulation and the real world using VR input devices.

\subsection{Software Engineering}


The following section contains some notes regarding the implementation of \robotio in order to explain some of the design choices.

\robotio is implemented in \textbf{python} a language commonly used for robotics, machine learning, and computer vision research. This allows interfacing with these kinds of software with minimal effort. As hardware interfaces will often require complex software libraries to be installed the installation of \robotio makes use of \textbf{conditional dependencies} to allow selecting those modules required for a certain machine. A more detailed description is given in Appendix \ref{sec:modules}.

Programming robots is difficult and doing so with \textbf{concurrency} doubly so. By default \robotio is designed to be used synchronously. Internally however, for several high latency task concurrency is used below the surface to speed up processes, an example of this is the \texttt{ThreadedCamera} class. Beyond this robot motion can be controlled asynchronously by setting the blocking keyword to \texttt{False} in actions. Finally, \texttt{async} versions of several functions are provided on an experimental basis.


The use of a configuration management system such as \textbf{hydra} \cite{Yadan2019} is a good match for robotics software, as seen in \cite{Murali2019}.
This allows keeping a structured overview over a large number of configuration parameters, makes it easy to compose configurations over sub-configurations, and also allows easy logging of the run-time configuration for better reproducibility of experiments.

Additional notes regarding convention on coordinate systems and the representation of geometric calculations are provided in Appendix \ref{sec:conventions}.

\subsection{Tuning Control}

Controlling a robot arm by streaming incremental actions is more complex than planning entire trajectories. For incremental actions the frequency and distances given as input must match the settings of the controller, otherwise the motion can be too slow or even jittery. These problems can be addressed by adjusting the control parameters of the underlying robot controller which can be found in the robot interface classes. We have tried to set sensible default values but would like to point out that these should be adjusted in necessary.

\vspace{.5em}
\section{Example Use-Cases}


Over the course of its development \robotio has been used and tested on a number of different robot setups that have been used for papers in several different areas of robotics, including reinforcement learning \cite{hermann2020, Nematollahi2022}, manipulation~\cite{von2022self, chisari2022correct}, learning from language and play \cite{Borja2022, Mees2022, hulc2022}, physics prediction \cite{Nematoli2020}, as well as visual servoing \cite{Argus2020, Izquierdo20}. Examples of this can be seen in Figure~\ref{fig:examples}.

\section{Discussion}

We present \robotio, a library for fast testing of real-world robot software. It provides common interfaces for robots allowing portable code to be written and programs to be tested on new setups with minimal adjustment.

We hope that sharing this work will allow more researchers to conduct real-world robot experiments, to spend less time on setting up robot systems, and more time developing new algorithms. We welcome additional contributions to extend \robotio.

\section*{Acknowledgments}

Lukas Hermann was the main developer of \robotio. This report was written by Max Argus, Adrian Röfer. Additional contributions to the code were made by Max Argus, Adrian Röfer, Iman Nematollahi, Oier Mees, Jessica Borja-Diaz, and Erick Rosete-Beas.

This work was partially funded by the German Research Foundation (DFG) under project numbers: 401269959 \& 417962828 and the German Federal Ministry of Education and Research under contract 01IS18040B-OML.

\section{Appendix}

\subsection{Optional Modules}
\label{sec:modules}
As hardware interfaces will often require installing complex software libraries, \robotio makes use of conditional dependencies to allow selecting only those modules required for a certain machine. A list of these modules, together with some of their requirements is provided in Table \ref{tab:modules}.

\begin{table}[!h]
    \centering
\begin{tabular}{|c|c|c|c|}
    \hline
    module & pip& kernel & rt \\
     & install & module &kernel  \\
    \hline
    ur{\_}rdte & \checkmark & & \\
    panda{\_}frankx & &  & \checkmark \\ 
    panda{\_}ros & & & \checkmark \\ 
    iiwa & & &\\ 
    realsese & & \checkmark &\\
    kinect & & &\\
    \hline
\end{tabular}
    \caption{A list of system requirements. pip install indicates if installation is possible using only pip, kernel modules indicate if the module requires installing kernel modules, and rt kernel indicates if the module requires running on a real-time kernel. 
    }
    \label{tab:modules}
\end{table}

\subsection{Conventions for Coordinate Systems and Units}
\label{sec:conventions}

Adhering to conventions as in the use of coordinate systems and units is important to ensure the compatibility and maintainability of code. For this project, we have specified the following conventions.

\begin{enumerate}
    \item Use a right-handed coordinate system
    \item Multiply transformations from the left $T_2 \times T_1$
    \item Use the quaternion order $(x, y, z, w)$. This is the convention used by bullet and scipy. Many parts of the code will detect quaternions from the numpy-quaternion package and convert these.
    \item The camera coordinate system on image: +x is right, +y is down, and +z is far.
    \item The TCP coordinate system should be: +x is right, +y is down, and +z is far.
    \item All distances and sizes should be in meters.
\end{enumerate}

\vspace{1em}

\FloatBarrier
\clearpage
\newpage

\bibliographystyle{IEEEtran}
\bibliography{references}

\begin{thebibliography}{10}
\providecommand{\url}[1]{#1}
\csname url@rmstyle\endcsname
\providecommand{\newblock}{\relax}
\providecommand{\bibinfo}[2]{#2}
\providecommand\BIBentrySTDinterwordspacing{\spaceskip=0pt\relax}
\providecommand\BIBentryALTinterwordstretchfactor{4}
\providecommand\BIBentryALTinterwordspacing{\spaceskip=\fontdimen2\font plus
\BIBentryALTinterwordstretchfactor\fontdimen3\font minus
  \fontdimen4\font\relax}
\providecommand\BIBforeignlanguage[2]{{%
\expandafter\ifx\csname l@#1\endcsname\relax
\typeout{** WARNING: IEEEtran.bst: No hyphenation pattern has been}%
\typeout{** loaded for the language `#1'. Using the pattern for}%
\typeout{** the default language instead.}%
\else
\language=\csname l@#1\endcsname
\fi
#2}}

\bibitem{kuka2016osmanual}
K.~R. GmbH, ``Kuka sunrise.os 1.11; kuka sunrise.workbench 1.11; operating and
  programming instructions for system integrators,'' Pub KUKA Sunrise.OS 1.11
  SI (PDF) en, 2016.

\bibitem{libfranka}
F.~Walch \emph{et~al.}, ``libfranka,''
  \url{https://github.com/frankaemika/libfranka}, accessed: 2022-07-01.

\bibitem{ur-rdte}
A.~P. Lindvig \emph{et~al.}, ``ur{\_}rtde,''
  \url{https://gitlab.com/sdurobotics/ur_rtde}, accessed: 2022-07-01.

\bibitem{Murali2019}
A.~Murali, T.~Chen, K.~V. Alwala, D.~Gandhi, L.~Pinto, S.~Gupta, and A.~K.
  Gupta, ``Pyrobot: An open-source robotics framework for research and
  benchmarking,'' \emph{ArXiv}, vol. abs/1906.08236, 2019.

\bibitem{Kroeger2011}
T.~Kröger, ``Opening the door to new sensor-based robot applications—the
  reflexxes motion libraries,'' in \emph{2011 IEEE International Conference on
  Robotics and Automation}, 2011, pp. 1--4.

\bibitem{Berscheid2021}
L.~Berscheid and T.~Kroeger, ``Jerk-limited real-time trajectory generation
  with arbitrary target states,'' in \emph{Robotics: Science and Systems XVII.
  Ed:: Dylan A. Shell}.\hskip 1em plus 0.5em minus 0.4em\relax {Robotics:
  Science and Systems Foundation}, 2021.

\bibitem{Coleman14}
D.~Coleman, I.~Sucan, S.~Chitta, and N.~Correll, ``Reducing the barrier to
  entry of complex robotic software: a moveit! case study,'' 04 2014.

\bibitem{sucan2012ompl}
I.~A. {\c{S}}ucan, M.~Moll, and L.~E. Kavraki, ``The {O}pen {M}otion {P}lanning
  {L}ibrary,'' \emph{{IEEE} Robotics \& Automation Magazine}, vol.~19, no.~4,
  pp. 72--82, December 2012.

\bibitem{kingston2019exploring}
Z.~Kingston, M.~Moll, and L.~E. Kavraki, ``Exploring implicit spaces for
  constrained sampling-based planning,'' \emph{Intl.\ J.\ of Robotics
  Research}, vol.~38, no. 10--11, pp. 1151--1178, Sept. 2019.

\bibitem{Kingston2021}
Z.~K. Kingston and L.~E. Kavraki, ``Robowflex: Robot motion planning with
  moveit made easy,'' \emph{ArXiv}, vol. abs/2103.12826, 2021.

\bibitem{Toussaint2014newton}
M.~Toussaint, ``Newton methods for k-order markov constrained motion
  problems,'' \emph{arXiv preprint arXiv:1407.0414}, 2014.

\bibitem{toussaint2015logic}
------, ``Logic-geometric programming: An optimization-based approach to
  combined task and motion planning,'' in \emph{Twenty-Fourth International
  Joint Conference on Artificial Intelligence}, 2015.

\bibitem{schulman2014motion}
J.~Schulman, Y.~Duan, J.~Ho, A.~Lee, I.~Awwal, H.~Bradlow, J.~Pan, S.~Patil,
  K.~Goldberg, and P.~Abbeel, ``Motion planning with sequential convex
  optimization and convex collision checking,'' \emph{The International Journal
  of Robotics Research}, vol.~33, no.~9, pp. 1251--1270, 2014.

\bibitem{Diankov2008}
R.~Diankov and J.~J. Kuffner, ``Openrave: A planning architecture for
  autonomous robotics,'' 2008.

\bibitem{Koenig2004}
N.~P. Koenig and A.~Howard, ``Design and use paradigms for gazebo, an
  open-source multi-robot simulator,'' \emph{2004 IEEE/RSJ International
  Conference on Intelligent Robots and Systems (IROS) (IEEE Cat.
  No.04CH37566)}, vol.~3, pp. 2149--2154 vol.3, 2004.

\bibitem{Coumans2021}
E.~Coumans and Y.~Bai, ``Pybullet, a python module for physics simulation for
  games, robotics and machine learning,'' \url{http://pybullet.org},
  2016--2021.

\bibitem{Todorov2012}
E.~Todorov, T.~Erez, and Y.~Tassa, ``Mujoco: A physics engine for model-based
  control,'' \emph{2012 IEEE/RSJ International Conference on Intelligent Robots
  and Systems}, pp. 5026--5033, 2012.

\bibitem{brax2021github}
\BIBentryALTinterwordspacing
C.~D. Freeman, E.~Frey, A.~Raichuk, S.~Girgin, I.~Mordatch, and O.~Bachem,
  ``Brax - a differentiable physics engine for large scale rigid body
  simulation,'' 2021. [Online]. Available: \url{http://github.com/google/brax}
\BIBentrySTDinterwordspacing

\bibitem{unity_robotics}
``Unity robotics,''
  \url{https://github.com/Unity-Technologies/Unity-Robotics-Hub}, accessed:
  2022-07-01.

\bibitem{Makoviychuk2021}
V.~Makoviychuk, L.~Wawrzyniak, Y.~Guo, M.~Lu, K.~Storey, M.~Macklin,
  D.~Hoeller, N.~Rudin, A.~Allshire, A.~Handa, and G.~State, ``Isaac gym: High
  performance gpu-based physics simulation for robot learning,'' \emph{ArXiv},
  vol. abs/2108.10470, 2021.

\bibitem{Brockman2016}
G.~Brockman, V.~Cheung, L.~Pettersson, J.~Schneider, J.~Schulman, J.~Tang, and
  W.~Zaremba, ``Openai gym,'' 2016.

\bibitem{Tassa2018}
Y.~Tassa, Y.~Doron, A.~Muldal, T.~Erez, Y.~Li, D.~de~Las~Casas, D.~Budden,
  A.~Abdolmaleki, J.~Merel, A.~Lefrancq, T.~P. Lillicrap, and M.~A. Riedmiller,
  ``Deepmind control suite,'' \emph{ArXiv}, vol. abs/1801.00690, 2018.

\bibitem{Szot2021}
A.~Szot, A.~Clegg, E.~Undersander, E.~Wijmans, Y.~Zhao, J.~Turner, N.~Maestre,
  M.~Mukadam, D.~S. Chaplot, O.~Maksymets, A.~Gokaslan, V.~Vondrus, S.~Dharur,
  F.~Meier, W.~Galuba, A.~X. Chang, Z.~Kira, V.~Koltun, J.~Malik, M.~Savva, and
  D.~Batra, ``Habitat 2.0: Training home assistants to rearrange their
  habitat,'' \emph{ArXiv}, vol. abs/2106.14405, 2021.

\bibitem{Kolve2017}
E.~Kolve, R.~Mottaghi, W.~Han, E.~VanderBilt, L.~Weihs, A.~Herrasti, D.~Gordon,
  Y.~Zhu, A.~K. Gupta, and A.~Farhadi, ``Ai2-thor: An interactive 3d
  environment for visual ai,'' \emph{ArXiv}, vol. abs/1712.05474, 2017.

\bibitem{microsoft_robotic_studio}
S.~Morgan, ``Simulating the world with microsoft robotics studio,''
  \url{https://docs.microsoft.com/en-us/archive/msdn-magazine/2008/june/robotics-simulating-the-world-with-microsoft-robotics-studio},
  accessed: 2022-07-01.

\bibitem{dfkirock}
T.~Röhr, ``Rock - the robot construction kit,''
  \url{https://robotik.dfki-bremen.de/en/research/softwaretools/rock.html},
  2013--2022.

\bibitem{Quigley2009}
M.~Quigley, ``Ros: an open-source robot operating system,'' in \emph{ICRA
  2009}, 2009.

\bibitem{Nematoli2020}
I.~Nematollahi, O.~Mees, L.~Hermann, and W.~Burgard, ``Hindsight for foresight:
  Unsupervised structured dynamics models from physical interaction,'' in
  \emph{Proceedings of the IEEE/RSJ International Conference on Intelligent
  Robots and Systems (IROS)}, Las Vegas, USA, 2020.

\bibitem{Borja2022}
J.~Borja-Diaz, O.~Mees, G.~Kalweit, L.~Hermann, J.~Boedecker, and W.~Burgard,
  ``Affordance learning from play for sample-efficient policy learning,'' in
  \emph{Proceedings of the IEEE International Conference on Robotics and
  Automation (ICRA)}, Philadelphia, USA, 2022.

\bibitem{Olson2011}
E.~Olson, ``{AprilTag}: A robust and flexible visual fiducial system,'' in
  \emph{Proceedings of the {IEEE} International Conference on Robotics and
  Automation ({ICRA})}.\hskip 1em plus 0.5em minus 0.4em\relax IEEE, May 2011,
  pp. 3400--3407.

\bibitem{Freipose2020}
\BIBentryALTinterwordspacing
C.~Zimmermann, A.~Schneider, M.~Alyahyay, T.~Brox, and I.~Diester, ``Freipose:
  A deep learning framework for precise animal motion capture in 3d spaces,''
  Tech. Rep., 2020. [Online]. Available:
  \url{https://lmb.informatik.uni-freiburg.de/projects/freipose/}
\BIBentrySTDinterwordspacing

\bibitem{yuan2017gelsight}
W.~Yuan, S.~Dong, and E.~H. Adelson, ``Gelsight: High-resolution robot tactile
  sensors for estimating geometry and force,'' \emph{Sensors}, vol.~17, no.~12,
  p. 2762, 2017.

\bibitem{Lambeta2020}
M.~Lambeta, P.~wei Chou, S.~Tian, B.~Yang, B.~Maloon, V.~R. Most, D.~Stroud,
  R.~Santos, A.~Byagowi, G.~Kammerer, D.~Jayaraman, and R.~Calandra, ``Digit: A
  novel design for a low-cost compact high-resolution tactile sensor with
  application to in-hand manipulation,'' \emph{IEEE Robotics and Automation
  Letters}, vol.~5, pp. 3838--3845, 2020.

\bibitem{bhirangi2021reskin}
R.~Bhirangi, T.~Hellebrekers, C.~Majidi, and A.~Gupta, ``Reskin:versatile,
  replaceable, lasting tactile skins,'' in \emph{CoRL}, 2021.

\bibitem{Mees2022}
O.~Mees, L.~Hermann, E.~Rosete-Beas, and W.~Burgard, ``Calvin - a benchmark for
  language-conditioned policy learning for long-horizon robot manipulation
  tasks,'' \emph{IEEE Robotics and Automation Letters (RA-L)}, vol.~7, no.~3,
  pp. 7327--7334, 2022.

\bibitem{Yadan2019}
\BIBentryALTinterwordspacing
O.~Yadan, ``Hydra - a framework for elegantly configuring complex
  applications,'' Github, 2019. [Online]. Available:
  \url{https://github.com/facebookresearch/hydra}
\BIBentrySTDinterwordspacing

\bibitem{hermann2020}
L.~Hermann, M.~Argus, A.~Eitel, A.~Amiranashvili, W.~Burgard, and T.~Brox,
  ``Adaptive curriculum generation from demonstrations for sim-to-real
  visuomotor control,'' in \emph{Proceedings of the International Conference on
  Robotics and Automation (ICRA)}, Paris, France, 2020.

\bibitem{Nematollahi2022}
I.~Nematollahi, E.~Rosete-Beas, A.~Rofer, T.~Welschehold, A.~Valada, and
  W.~Burgard, ``Robot skill adaptation via soft actor-critic gaussian mixture
  models,'' \emph{2022 International Conference on Robotics and Automation
  (ICRA)}, pp. 8651--8657, 05 2022.

\bibitem{von2022self}
J.~O. von Hartz, E.~Chisari, T.~Welschehold, and A.~Valada, ``Self-supervised
  learning of multi-object keypoints for robotic manipulation,'' \emph{arXiv
  preprint arXiv:2205.08316}, 2022.

\bibitem{chisari2022correct}
E.~Chisari, T.~Welschehold, J.~Boedecker, W.~Burgard, and A.~Valada, ``Correct
  me if i am wrong: Interactive learning for robotic manipulation,'' \emph{IEEE
  Robotics and Automation Letters}, vol.~7, no.~2, pp. 3695--3702, 2022.

\bibitem{hulc2022}
O.~Mees, L.~Hermann, and W.~Burgard, ``What matters in language conditioned
  robotic imitation learning,'' \emph{arXiv preprint arXiv:2204.06252}, 2022.

\bibitem{Argus2020}
M.~Argus, L.~Hermann, J.~Long, and T.~Brox, ``Flowcontrol: Optical flow based
  visual servoing,'' in \emph{Proceedings of the International Conference on
  Intelligent Robots and System (IROS)}, Las Vegas, Arizona, 2020.

\bibitem{Izquierdo20}
S.~Izquierdo, M.~Argus, and T.~Brox, ``Conditional visual servoing for
  multi-step tasks,'' in \emph{Proceedings of the IEEE/RSJ International
  Conference on Intelligent Robots and Systems (IROS)}, 2022.

\end{thebibliography}

\end{document}